\documentclass{article}

\usepackage[accepted]{icml2014}

\usepackage{graphicx} 
\usepackage{subfigure}
\usepackage{color}
\definecolor{darkblue}{rgb}{0,0,0.5}
\definecolor{firebrick}{rgb}{0.75,0.125,0.125}
\definecolor{darkgreen}{rgb}{0,0.5,0}
\definecolor{light-gray}{gray}{0.5}
\usepackage[colorlinks=true,linkcolor=firebrick,citecolor=darkgreen,urlcolor=darkblue]{hyperref}

\usepackage{amsthm} 
\usepackage{latexsym,amsmath,amsfonts,amssymb} 
\usepackage{url}
\usepackage{graphics, color}
\usepackage{graphicx}
\usepackage{fullpage}
\usepackage{notations}
\usepackage{myAlgorithm}

\usepackage%
{trackchanges}
\renewcommand{\initialsTwo}{BK}
\renewcommand{\initialsThree}{RBF}

\icmltitlerunning{}

\allowdisplaybreaks[4]

\begin{document} 

\twocolumn[
\icmltitle{Correlation-based construction of neighborhood and edge features}

\icmlauthor{Bal\'azs K\'egl}{balazs.kegl@gmail.com}
\icmladdress{LAL/LRI, University of Paris-Sud, CNRS, 91898
Orsay, France}

\icmlkeywords{}

\vskip 0.3in
]

\begin{abstract}
Motivated by an abstract notion of low-level edge detector filters, we propose a
simple method of unsupervised feature construction based on pairwise statistics
of features. In the first step, we construct neighborhoods of features by
regrouping features that correlate. Then we use these subsets as filters to
produce new neighborhood features. Next, we connect neighborhood features that
correlate, and construct edge features by subtracting the correlated
neighborhood features of each other. To validate the usefulness of the
constructed features, we ran AdaBoost.MH on four multi-class classification
problems. Our most significant result is a test error of $0.94\%$ on MNIST with
an algorithm which is essentially free of any image-specific priors. On CIFAR-10
our method is suboptimal compared to today's best deep learning techniques,
nevertheless, we show that the proposed method outperforms not only boosting on
the raw pixels, but also boosting on Haar filters.
\end{abstract}

\section{Introduction}

In this paper we propose a simple method of unsupervised feature construction
based on pairwise statistics of features. In the first step, we construct
neighborhoods of features by regrouping features that correlate. Then we use
these subsets of features as filters to produce new \emph{neighborhood
  features}. Next, we connect neighborhood features that correlate, and
construct \emph{edge features} by subtracting the correlated neighborhood
features of each other. The method was motivated directly by the notion of
low-level edge detector filters. These filters work well in practice, and they
are ubiquitous in the first layer of both biological and artificial systems that
learn on natural images. Indeed, the four simple feature construction steps are
directly based on an abstract notion of Haar or Gabor filters: homogeneous,
locally connected patches of contrasting intensities. In a more high-level
sense, the technique is also inspired by a, perhaps naive, notion of how natural
neural networks work: in the first layer they pick up correlations in stimuli,
and settle in a ``simple'' theory of the world. Next they pick up events when
the correlations are broken, and assign a new units to the new ``edge''
features. Yet another direct motivation comes from \cite{LBLJK07} where they
show that pixel order can be recovered using only feature correlations. Once
pixel order is recovered, one can immediately apply algorithms that explicitly
use neighborhood-based filters. From this point of view, in this paper we show
that it is possible to go from pixel correlations to feature construction,
without going through the explicit mapping of the pixel order.

To validate the usefulness of the constructed features, we ran
\Algo{AdaBoost.MH} \cite{ScSi99} on multi-class classification
problems. Boosting is one of the best ``shallow'' multi-class classifiers,
especially when the goal is to combine simple classifiers that act on small
subsets a large set of possibly useful features. Within multi-class boosting
algorithms, \Algo{AdaBoost.MH} is the state of the art. On this statement we are
at odds with recent results on multiclass boosting. On the two UCI data sets we
use in this paper (and on others reported in \cite{KeBu09}), \Algo{AdaBoost.MH}
with Hamming trees and products \cite{BBCCK11} clearly outperforms \Algo{SAMME}
\cite{ZhZoRoHa09}, \Algo{ABC-Boost} \cite{Li09}, and most importantly, the
implementation of \Algo{AdaBoost.MH} in \cite{ZhZoRoHa09}, suggesting that
\Algo{SAMME} was compared to a suboptimal implementation of \Algo{AdaBoost.MH}
in \cite{ZhZoRoHa09}.

Our most significant result comes on the MNIST set where we achieve a test error
of $0.94\%$ with an algorithm which is essentially free of any image-specific
priors (e.g., pixel order, preprocessing). On CIFAR-10, our method is suboptimal
compared to today's best deep learning techniques, reproducing basically the
results of the earliest attempts on the data set \cite{RaKiHi10}. The main point
in these experiments is that the proposed method outperforms not only boosting
on the raw pixels, but also boosting on Haar filters. We also tried the
technique on two relatively large UCI data sets. The results here are
essentially negative: the lack of correlations between the features do not allow
us to improve significantly on ``shallow'' \Algo{AdaBoost}.

The paper is organized as follows. In Section~\ref{secConstructing} we formally
describe the method. In Section~\ref{secExperiments} we show experimental
results, and we conclude with a discussion on future research in
Section~\ref{secConclusions}.

\section{Constructing the representation}
\label{secConstructing}

For the formal description of the method, let $\cD =
\big\{(\bx_1,y_1),\ldots,(\bx_n,y_n)\big\}$ the training data, where $\bx_i =
(x_i^1,\ldots,x_i^d) \in \RR^d$ are the input vectors and $y_i \in
\{1,\ldots,K\}$ the labels. We will denote the input matrix and its elements by
$\cX = [x_i^j]_{i=1,\ldots,n}^{j=1,\ldots,d}$, and its raw feature (column) vectors
by $\bx^j = (x_1^j,\ldots,x_n^j) \in \RR^n$. The algorithm consists of the
following steps.
\begin{enumerate}
\item We construct \emph{neighborhoods} $J(\bx^j)$ for each feature vector
  $\bx^j$, that is, sets of features\footnote{In what follows, we use the word
  \emph{feature} for any real-valued representation of the input, so both the
  input of the feature-building algorithm and its ouput. It is an intended
  terminology to emphasize that the procedure can be applied recursively, as in
  stacked autoencoders.} that are correlated with $\bx^j$. Formally, $J(\bx^j) =
  \{j^\prime:\bx^{j^\prime} \in \cX: \rho(\bx^j,\bx^{j^\prime}) <
  \rho_{\text{N}}\}$, where $$\rho(\bx^j,\bx^{j^\prime}) =
  \frac{\COV{\bx^j,\bx^{j^\prime}}}{\sqrt{\VAR{\bx^j}\VAR{\bx^{j^\prime}}}}$$ is
  the correlation of two feature vectors and $\rho_{\text{N}}$ is a
  hyperparameter of the algorithm.
\item We construct \emph{neighborhood features} by using the neighborhoods as
  (normalized) filters, that is, $$z_i^j = \frac{1}{|J(\bx^j)|}\sum_{j^\prime \in
  J(\bx^{j})} x_i^{j^\prime}$$ for all $i=1,\ldots,n$ and $j=1\ldots,d$.
\item We construct \emph{edges} between neighborhoods by connecting correlated
  neighborhood features. Formally, $\cL = \{(j_1,j_2): 1\le j_1,j_2\le d, \;
  \rho(\bz^{j_1},\bz^{j_2}) < \rho_{\text{E}}\}$, where $\rho_{\text{E}}$ is a
  hyperparameter of the algorithm. We will denote elements of the set of edges
  by $(\ell_{j,1},\ell_{j,2})$, and the size of the set by $|\cL| = L$.
\item We construct \emph{edge features} by \emph{subtracting} the responses to
  correlated neighborhoods of each other, that is, $s_i^j = z_i^{\ell_{j,1}} -
  z_i^{\ell_{j,2}}$ for all $j=1,\ldots,L$ and $(\ell_{j,1},\ell_{j,2}) \in
  \cL$.
\item We concatenate neighborhood and edge features into a new representation of
  $\bx_i$, that is, $\bx^\prime_i =
  \big(z_i^1,\ldots,z_i^d,s_i^1,\ldots,s_i^L\big)$ for all $i=1,\ldots,n$.

\end{enumerate}

\subsection{Setting $\rho_{\text{N}}$ and $\rho_{\text{E}}$}

Both hyperparameters $\rho_{\text{N}}$ and $\rho_{\text{E}}$ threshold
correlations, nevertheless, they have quite different roles: $\rho_{\text{N}}$
controls the neighborhood size whereas $\rho_{\text{E}}$ controls the distance
of neighborhoods under which an edge (that is, a significantly different
response or a ``surprise'') is an interesting feature. In practice, we found
that the results were rather insensitive to the value of these parameters in the
$[0.3,0.9]$ interval. For now we manually set $\rho_{\text{N}}$ and
$\rho_{\text{E}}$ in order to control the number of features. In our experiments
we found that it was rarely detrimental to increase the number of features in
terms of the asymptotic test error (w.r.t. the number of boosting iterations
$T$) but the convergence of \Algo{AdaBoost.MH} slowed down if the number of
features were larger than some thousands (either because each boosting iteration
took too much time, or because we had to seriously subsample the features in
each iteration, essentially generating random trees, and so the number of
iterations exploded).

On images, where the dimensionality of the input space (number of pixels) is
large, we subsample the pixels $[\bx^j]_{j=1,\ldots,d}$ before constructing the
neighborhoods $J(\bx^j)$ to control the number of neighborhood features, again,
for computational rather than statistical reasons. We simply run
\Algo{AdaBoost.MH} with decision stumps in an autoassociative setup. A decision
stump uses a single pixel as input and outputs a prediction on all the
pixels. We take the first $d^\prime$ stumps that \Algo{AdaBoost} picks, and use
the corresponding pixels (a subset of the full set of pixels) to construct the
neighborhoods.

On small-dimensional (non-image) sets we face the opposite problem: the small
number of input features limit the number of neighborhoods. This actually
highlights a limitation of the algorithm: when the number of input features is
small and they are not very correlated, the number of generated neighborhood and
edge features is small, and they essentially contain the same information as the
original features. Nevertheless, we were curious whether we can see any
improvement by blowing up the number of features (similarly, in spirit, to what
support vector machines do). We obtain a larger number of neighborhoods by
defining a \emph{set} of thresholds
$\{\rho_{\text{N}}^1,\ldots,\rho_{\text{N}}^M\}$, and constructing $M$
``concentric'' neighborhoods for each feature $\bx^j$. On data sets with
heterogeneous feature types it is also important to normalize the features by
the usual transformation ${x^j_i}^\prime = \big(x^j_i -
\mu(\bx^j)\big)/\sigma(\bx^j)$, where $\mu(\bx)$ and $\sigma(\bx)$ denote the
mean and the standard deviation of the elements of $\bx$, respectively, before
proceeding with the feature construction (Step~2).

Optimally, of course, automatic hyperparameter optimization
\cite{BeBe12,BeBaKeBe11,SnLaAd12}
is the way to go, especially since \Algo{AdaBoost.MH} also has two or three
hyperparameters, and manual grid search in a four-to-five dimensional
hyperparameter space is not feasible. For now, we set aside this issue for
future work.

\subsection{\Algo{AdaBoost.MH} with Hamming trees}

The constructed features can be input to any ``shallow'' classifier. Since we
use \Algo{AdaBoost.MH} with Hamming trees, we briefly describe them here. The
full formal description with the pseudocode is in the documentation of
\Algo{MultiBoost} \cite{BBCCK11}. It is available at the
\href{http://multiboost.org}{\tt multiboost.org} website along with the code
itself.

The advantage of $\Algo{AdaBoost.MH}$ over other multi-class boosting approaches
is that it does not require from the base learner to predict a single label
$\widehat{y} = h(\bx)$ for an input instance $\bx$, rather, it uses
\emph{vector-valued} base learners $\bh(\bx)$. The requirement for these base
learners is weaker: it suffices if the edge $\gamma(\bh,\bW) = \sum_{i=1}^n
\sum_{\ell=1}^K w_{i, \ell} h_\ell(x_i)y_{i,\ell}$ is slightly larger than zero,
where $\bW = [w_{i, \ell}]_{i=1,\ldots,n}^{\ell=1,\ldots,K}$ is the weight
matrix (over instances and labels) in the current boosting iteration, and $\by_i
= (y_{i,1},\ldots,y_{i,K})$ is a $\pm 1$-valued one-hot code of the label. This
makes it easy to turn weak \emph{binary} classifiers into multi-class base
classifiers, without requiring that the multi-class \emph{zero-one} base error
be less than $1/2$. In case $\bh$ is a decision tree, there are two important
consequences. First, the size (the number of leaves $N$) of the tree can be
arbitrary and can be tuned freely (whereas requiring a zero-one error to be less
than $1/2$ usually implies large trees). Second, one can design trees with
binary $\{\pm 1\}^K$-valued outputs, which could not be used as standalone
multi-class classifiers.

In a Hamming tree, at each node, the split is learned by a multi-class decision
stump of the form $\bh(\bx) = \bv \varphi(\bx)$, where $\varphi(\bx)$ is a
(standard) scalar $\pm 1$-valued decision stump, and $\bv$ is a $\{\pm
1\}^K$-valued vector. At leaf nodes, the full $\{\pm 1\}^K$-valued vector $\bv
\varphi(\bx)$ is output for a given $\bx$, whereas at inner nodes, only the
binary function $\varphi(\bx)$ is used to decide whether the instance $\bx$ goes
left or right. The tree is constructed top-down, and each node stump is
optimized in a greedy manner, as usual. For a given $\bx$, unless $\bv$ happens
to be a one-hot vector, no single class can be output (because of ties). At the
same time, the tree is perfectly boostable: the weighted \emph{sum} of Hamming
trees produces a \emph{real}-valued vector of length $K$, of which the predicted
class can be easily derived using the $\argmax$ operator.

\section{Experiments}
\label{secExperiments}

We carried out experiments on four data sets:
MNIST\footnote{\url{http://yann.lecun.com/exdb/mnist}} and
CIFAR-10\footnote{\url{http://www.cs.toronto.edu/~kriz/cifar.html}} are standard
image classification data sets, and the Pendigit and Letter sets are relatively
large benchmarks from the UCI
repository\footnote{\url{http://www.ics.uci.edu/~mlearn/MLRepository.html}}. We
boosted Hamming trees \cite{BBCCK11} on each data sets. Boosted Hamming trees
have three hyperparameters: the number of boosting iterations $T$, the number of
leaves $N$, and the number of (random) features $d^\prime$ considered at each
split (\Algo{LazyBoost}~\cite{EsMaRi00} settings that give the flavor of a
random forest to the final classifier). Out of these three, we validated only
the number of leaves $N$ in the ``classical'' way using $80/20$ single
validation on the training set. Since \Algo{AdaBoost.MH} does not exhibit any
overfitting even after a very large number of iterations (see
Figure~\ref{figLearningCurves}),\footnote{This is certainly the case in these
  four experiments, but in our experience, making \Algo{AdaBoost.MH} overfit is
  really hard unless significant label noise is present in the training set.} we
run it for a large number of $T=10^5$ iterations, and report the average test
error of the last $T/2$ iterations. 
The number of (random) features $d^\prime$ considered at each split is another
hyperparameter which does not have to be tuned in the traditional way. In our
experience, the larger it is, the smaller the \emph{asymptotic} test error
is. On the other hand, the larger it is the slower the algorithm converges to
this error. This means that $d^\prime$ controls the trade-off between the
accuracy of the final classifier and the training time. We tuned it to obtain
the full learning curves in reasonable time.

The neighborhood and edge features were constructed as described in
Section~\ref{secConstructing}. Estimating correlations can be done robustly on
relatively small random samples, so we ran the algorithm on a random input
matrix $\cX$ with $1000$ instances.

\subsection{MNIST}

\begin{figure*}[!ht]
\centerline{
  \parbox{0.5\textwidth}{
    \subfigure[MNIST]{\label{figMNIST}
    \includegraphics[width=0.5\textwidth]{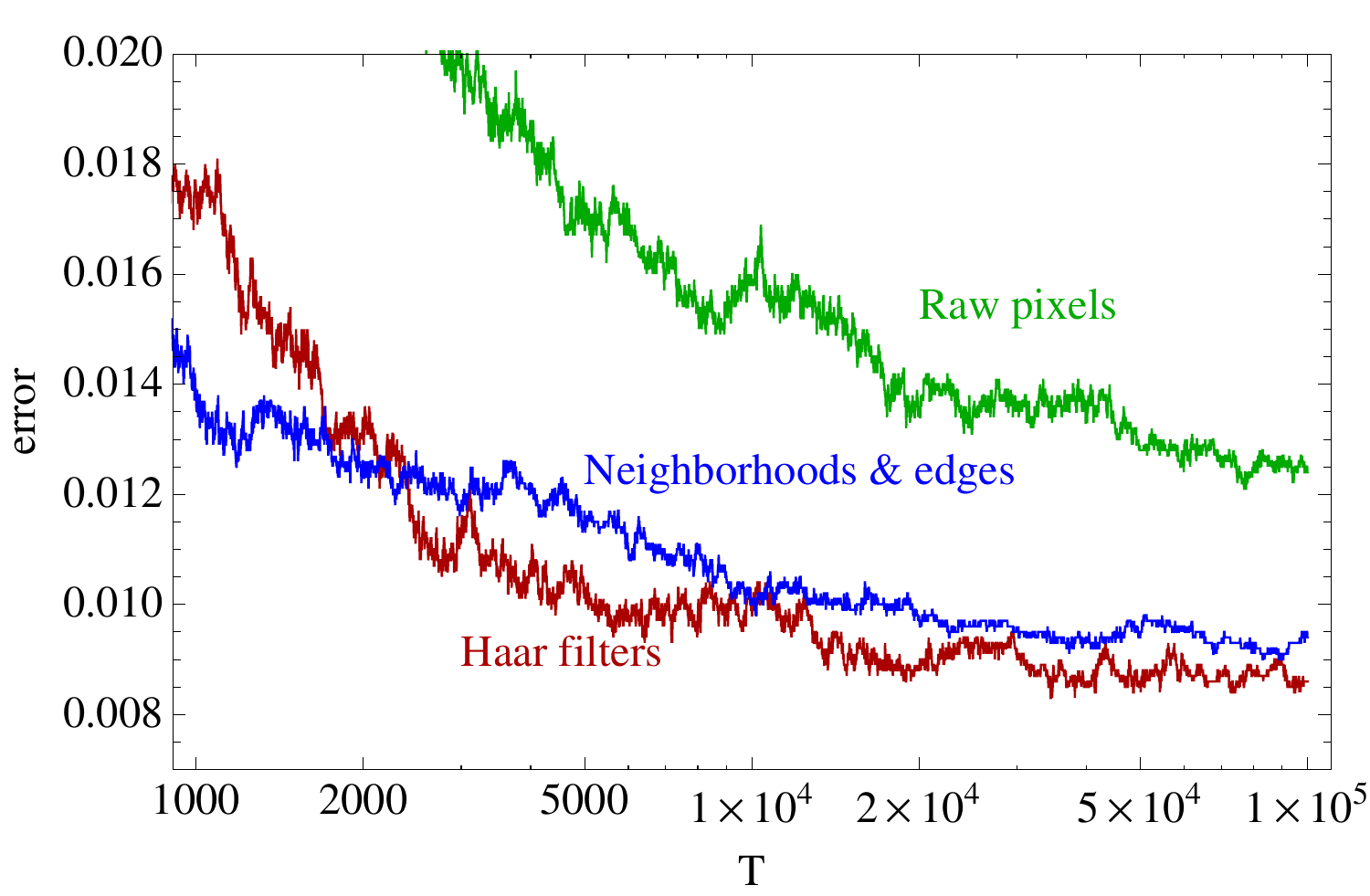}
    }
  }
  \parbox{0.5\textwidth}{
    \subfigure[CIFAR-10]{\label{figCIFAR}
    \includegraphics[width=0.5\textwidth]{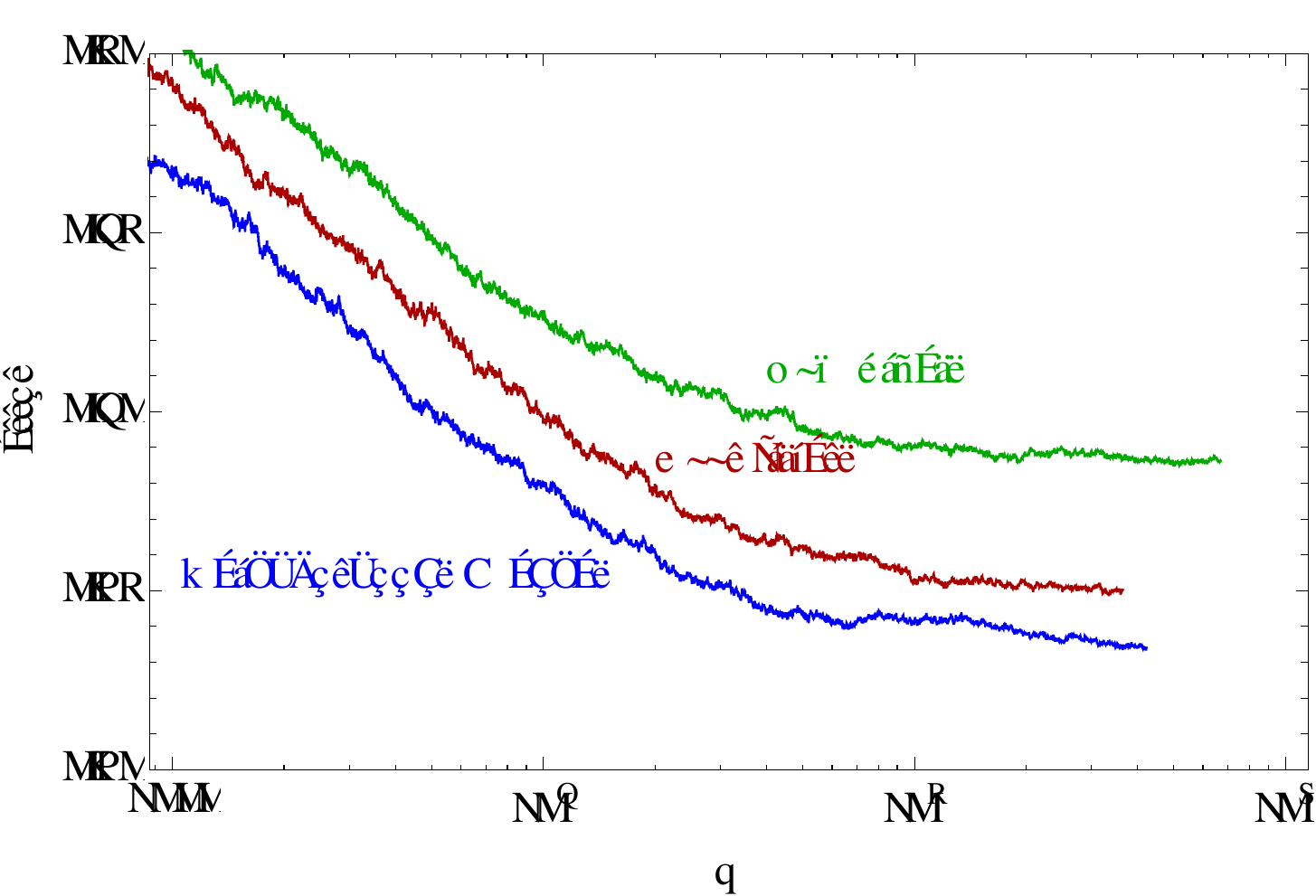}
    }
  }
}
\centerline{
  \parbox{0.5\textwidth}{
    \subfigure[UCI Letter]{\label{figLetter}
    \includegraphics[width=0.5\textwidth]{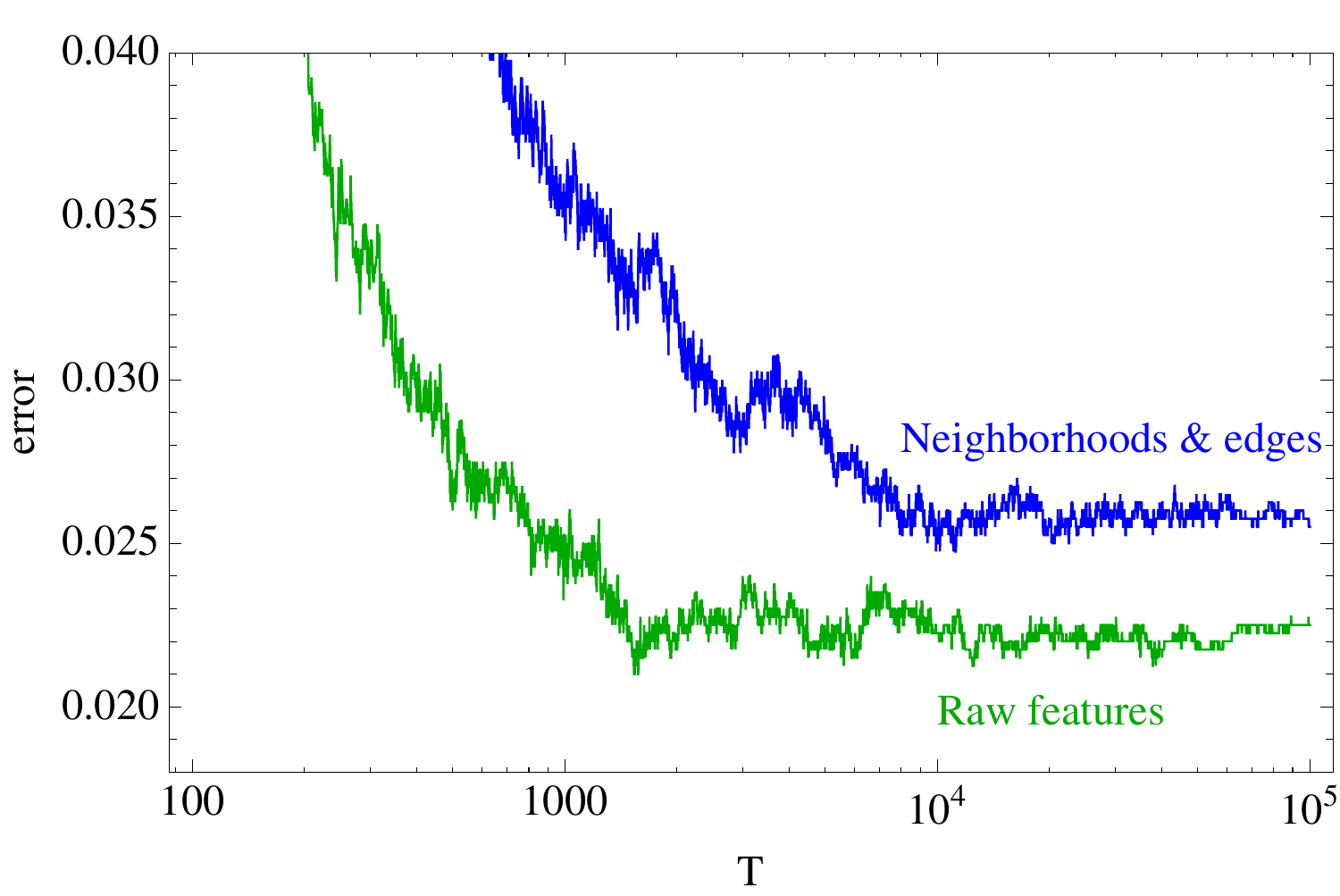}
    }
  }
  \parbox{0.5\textwidth}{
    \subfigure[UCI Pendigit]{\label{figPendigit}
    \includegraphics[width=0.5\textwidth]{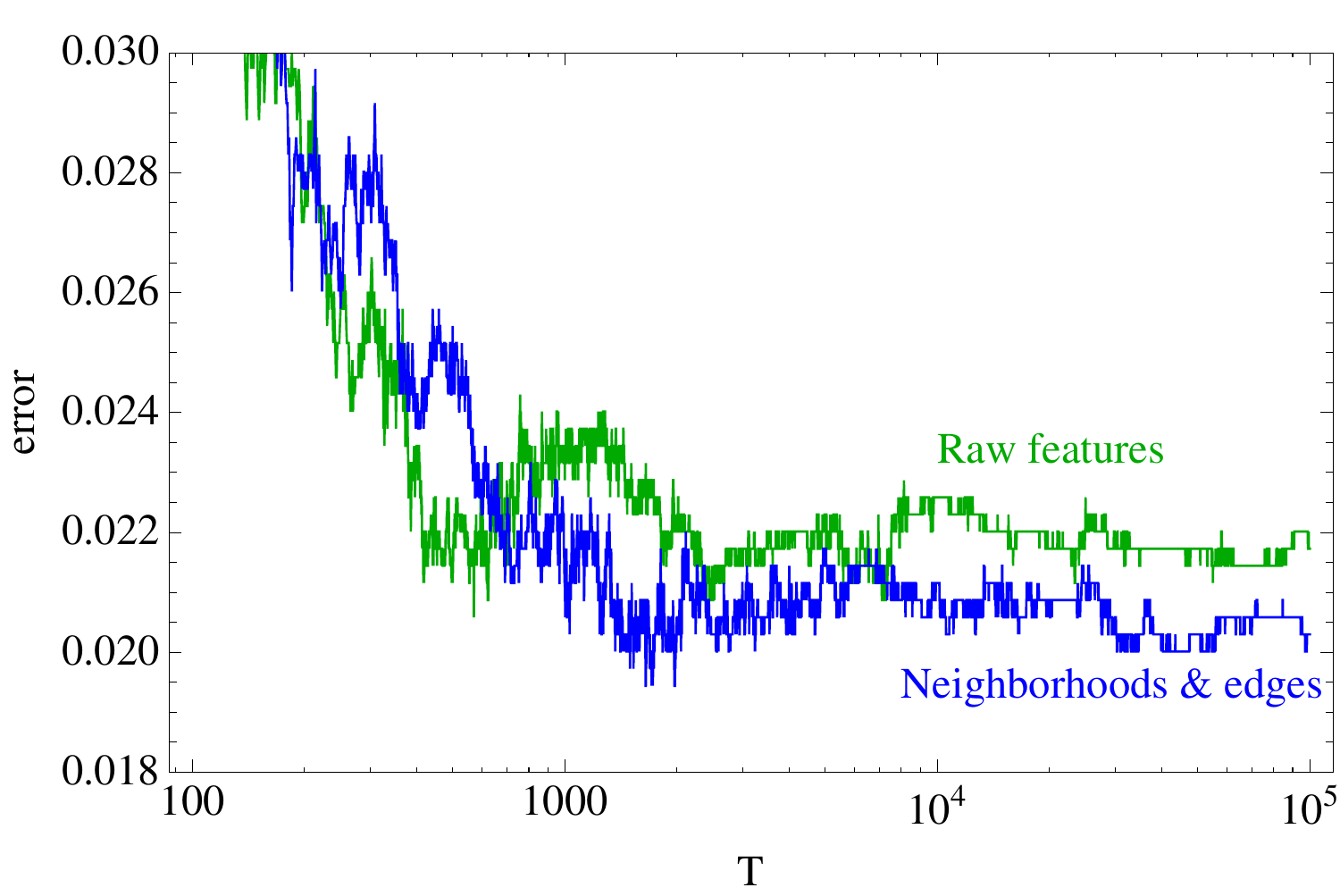}
    }
  }
}
\caption{Test learning curves. \label{figLearningCurves}}
\end{figure*}

\begin{figure*}[!ht]
\centerline{
  \parbox{0.35\textwidth}{
    \subfigure[MNIST]{\label{figMNISTPixels}
    \includegraphics[width=0.35\textwidth]{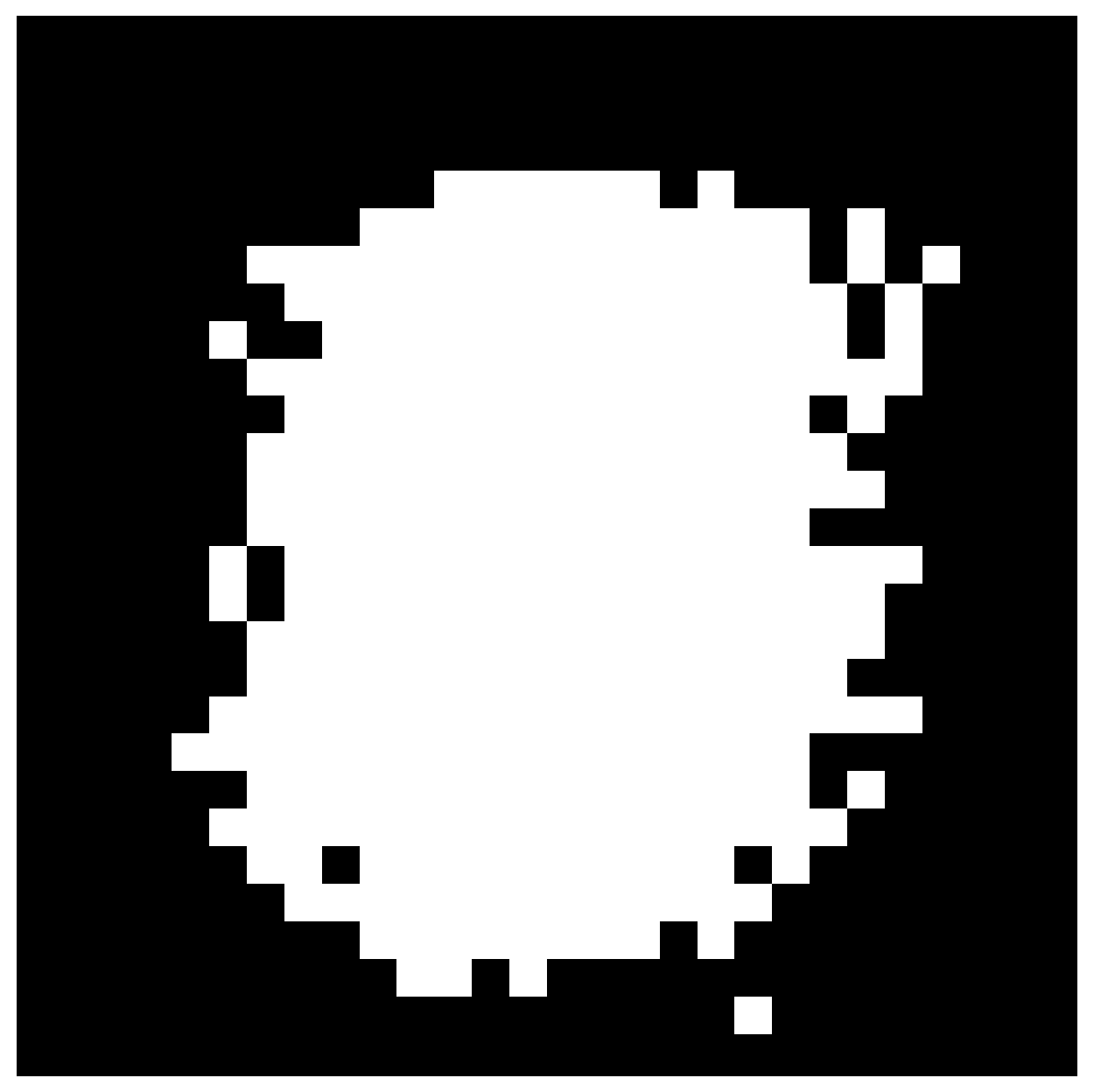}
    }
  }
  \parbox{0.3\textwidth}{
    \subfigure[CIFAR-10]{\label{figCIFARPixels}
    \includegraphics[width=0.35\textwidth]{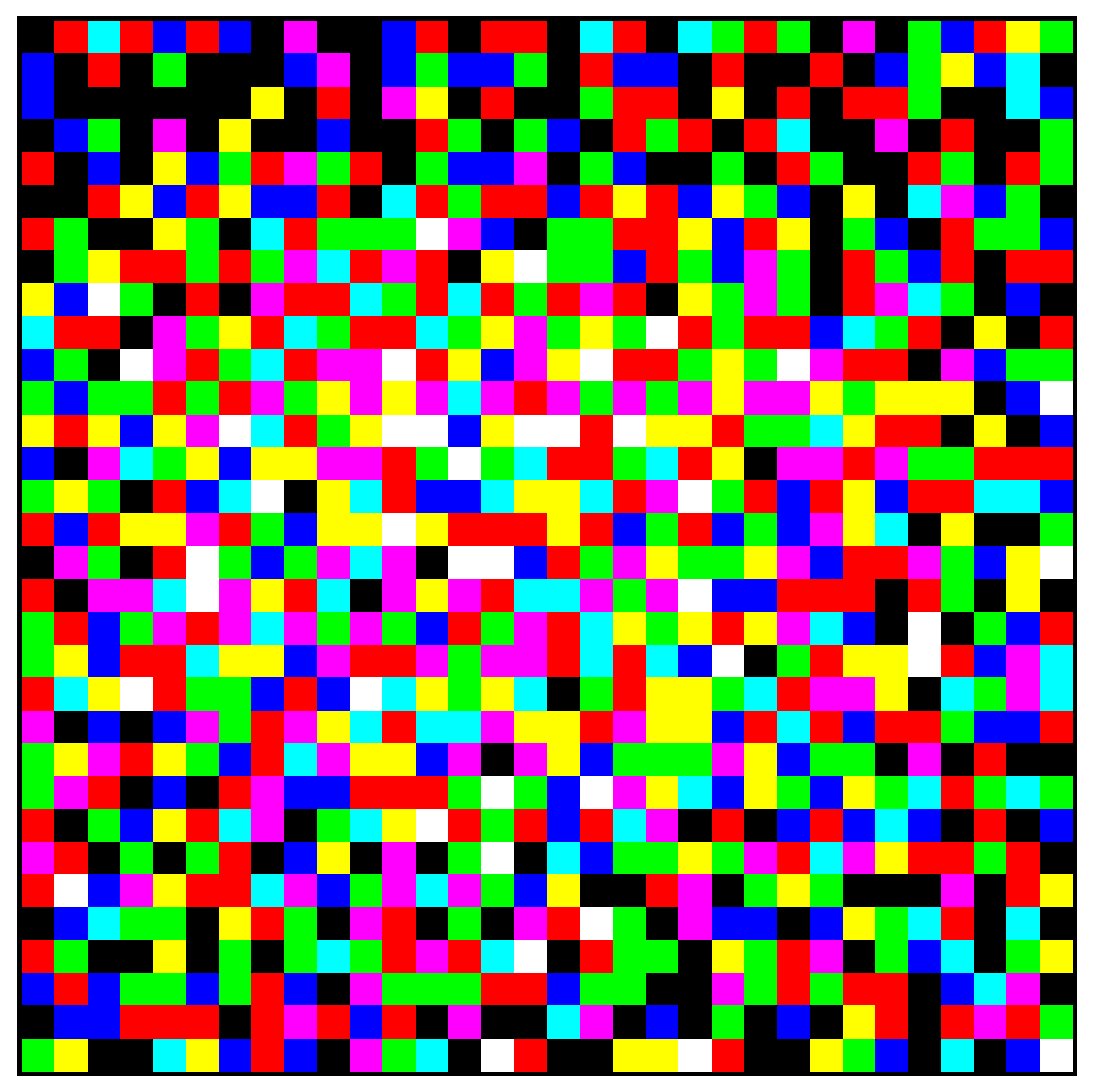}
    }
  }
}
\caption{Initial pixel selection using autoassociative \Algo{AdaBoost.MH}. (a)
  White means that the pixel is selected, black means it is not. (b) Colors are
  mixtures of the color channels selected. White means that all channels were
  selected, and black means that none of them were
  selected. \label{figSelectedPixels}}
\end{figure*}

\begin{figure*}[!ht]
\centerline{
  \parbox{1.0\textwidth}{
    \includegraphics[width=1.0\textwidth]{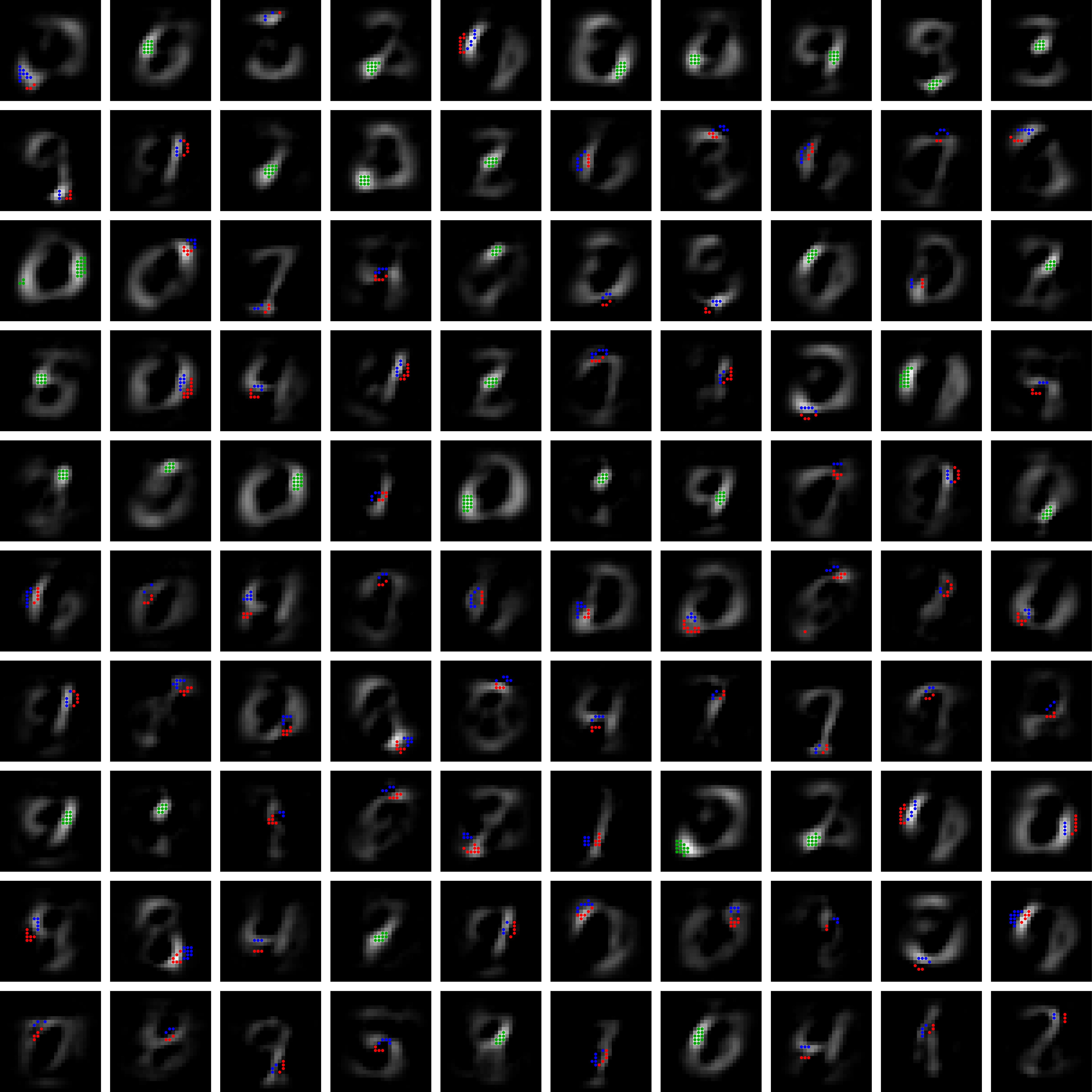}
  }
}
\caption{The $100$ most important neighborhood and edge features picked by
  \Algo{AdaBoost.MH}. In neighborhood features we mark the middles of pixels in
  the neighborhood $J(\bx^j)$ by green dots. In edge features we mark the
  middles of pixels of the positive neighborhood by blue dots, and the pixels of
  the negative neighborhood by red dots. The black\&white images in both cases
  are averages of the 20\% of MNIST test images that respond the strongest to
  the given filter. \label{figMNISTFilters}}
\end{figure*}

MNIST consists of $60000$ grey-scale training images of hand-written digits of
size $28\times 28 = 784$. In all experiments on MNIST, $d^\prime$ was set to
$100$. The first baseline run was \Algo{AdaBoost.MH} with Hamming trees of $8$
leaves on the raw pixels (green curve in Figure~\ref{figMNIST}), achieving a
test error of $1.25\%$. We also ran \Algo{AdaBoost.MH} with Hamming trees of $8$
leaves in the roughly $300000$-dimensional feature space generated by five types
of Haar filters (\citealt{ViJo01}; red curve in Figure~\ref{figMNIST}). This
setup produced a test error of $0.85\%$ which is the state of the art among
boosting algorithms. For generating neighborhood and edge features, we first ran
autoassociative \Algo{AdaBoost.MH} with decision stumps for $800$ iterations
that picked the $326$ pixels depicted by the white pixels in
Figure~\ref{figMNISTPixels}. Then we constructed $326$ neighborhood features
using $\rho_{\text{N}} = 0.5$ and $1517$ edge features using $\rho_{\text{E}} =
0.7$. The $100$ most important features (picked by running \Algo{AdaBoost.MH}
with decision stumps) is depicted in Figure~\ref{figMNISTFilters}. Finally we
ran \Algo{AdaBoost.MH} with Hamming trees of $8$ leaves on the constructed
features (blue curve in Figure~\ref{figMNIST}), achieving a test error of
$0.94\%$ which is one of the best results among methods that do not use explicit
image priors (pixel order, specific distortions, etc.).

Note that \Algo{AdaBoost.MH} picked slightly more neighborhood than edge
features relatively to their prior proportions. On the other hand, it was
crucial to include both neighborhood and edge features: \Algo{AdaBoost.MH} was
way suboptimal on either subset.

\subsection{CIFAR}

CIFAR-10 consists of $50000$ color training images of 10 object categories of
size $32\times 32 = 1024$, giving a total of $3\times 1024 = 3072$ features. In
all experiments on CIFAR, $d^\prime$ was set to $10$. The first baseline run was
\Algo{AdaBoost.MH} with Hamming trees of $20$ leaves on the raw pixels (green
curve in Figure~\ref{figCIFAR}), achieving a test error of $39.1\%$. We also ran
\Algo{AdaBoost.MH} with Hamming trees of $20$ leaves in the roughly
$350000$-dimensional feature space generated by five types of Haar filters
(\citealt{ViJo01}; red curve in Figure~\ref{figCIFAR}). This setup produced a
test error of $36.1\%$. For generating neighborhood and edge features, we first
ran autoassociative \Algo{AdaBoost.MH} with decision stumps for $1000$
iterations that picked the $922$ color channels depicted by the white and
colored pixels in Figure~\ref{figCIFARPixels}. Then we constructed $922$
neighborhood features using $\rho_{\text{N}} = 0.85$ and $4552$ edge features
using $\rho_{\text{E}} = 0.85$. Finally we ran \Algo{AdaBoost.MH} with Hamming
trees of $20$ leaves on the constructed features (blue curve in
Figure~\ref{figMNIST}), achieving a test error of $33.6\%$. 

None of these results are close to the sate of the art, but they are not
completely off the map, either: they match the performance of one of the early
techniques that reported error on CIFAR-10 \cite{RaKiHi10}. The main
significance of this experiment is that \Algo{AdaBoost.MH} with neighborhood and
edge features can beat not only \Algo{AdaBoost.MH} on raw pixels but also
\Algo{AdaBoost.MH} with Haar features.

\subsection{UCI Pendigit and Letter}

In principle, there is no reason why neighborhood and edge features could not
work on non-image sets. To investigate, we ran some preliminary tests on the
relatively large UCI data sets, Pendigit and Letter. Both of them contain $16$
features and several thousand instances. The baseline results are $2.16\%$ on
Pendigit using \Algo{AdaBoost.MH} with Hamming trees of $4$ leaves (green curve
in Figure~\ref{figPendigit}) and $2.23\%$ on Letter using \Algo{AdaBoost.MH}
with Hamming trees of $20$ leaves (green curve in Figure~\ref{figLetter}). We
constructed neighborhoods using a set of thresholds $\rho_{\text{N}} =
\{0.1,0.2,\ldots,0.9\}$, giving us $62$ unique neighborhoods on Letter and $86$
unique neighborhoods on Pendigit (out of the possible $9\times 16=126$). We then
proceeded by constructing edge features with $\rho_{\text{E}} = 0.7$, giving us
$506$ more features on Letter, and $1040$ more features on Letter. We then ran
\Algo{AdaBoost.MH} with Hamming trees of the same number of leaves as in the
baseline experiments, using $d^\prime=100$. On Pendigit, we obtained $2.05\%$,
better than in the baseline (blue curve in Figure~\ref{figPendigit}), while on
Letter we obtained $2.59\%$, significantly worse than in the baseline (blue
curve in Figure~\ref{figLetter}). We see two reasons why a larger gain is
difficult on these sets. First, there is not much correlation between the
features to exploit. Indeed, setting $\rho_{\text{N}}$ and $\rho_{\text{E}}$ to
similar values to those we used on the image sets, neighborhoods would have been
very small, and there would have been almost no edges. Second,
\Algo{AdaBoost.MH} with Hamming trees is already a very good algorithm on these
sets, so there is not much margin for improvement.

\section{Future work}
\label{secConclusions}

Besides running more experiments and tuning hyperparameters automatically, the
most interesting question is whether stacking neighborhood and edge features
would work. There is no technical problem of re-running the feature construction
on the features obtained in the first round, but it is not clear whether there
is any more structure this simple method can exploit. We did some preliminary
trials on MNIST where it did not improve the results, but this may be because
MNIST is a relatively simple set with not very complex features and rather
homogeneous classes. Experimenting with stacking on CIFAR is definitely the next
step. Another interesting avenue is to launch a large scale exploration on more
non-image benchmark sets to see whether there is a subclass of sets where the
correlation-based feature construction may work, and then to try to characterize
this subclass.


\end{document}